\documentclass[10pt,twocolumn,letterpaper]{article}

\usepackage[pagenumbers]{cvpr} 
\usepackage{pifont}

\usepackage{url}

\usepackage[utf8]{inputenc} 
\usepackage[T1]{fontenc}    
\usepackage{url}            
\usepackage{booktabs}       
\usepackage{amsfonts}       
\usepackage{nicefrac}       
\usepackage{microtype}      
\usepackage{xcolor}         

\usepackage[utf8]{inputenc} 
\usepackage[T1]{fontenc}    
\usepackage{url}            
\usepackage{booktabs}       
\usepackage{amsfonts}       
\usepackage{nicefrac}       
\usepackage{microtype}      
\usepackage{xcolor}         
\usepackage{natbib}
\PassOptionsToPackage{options}{natbib}

\usepackage{times}
\usepackage{soul}
\usepackage{url}
\usepackage[utf8]{inputenc}
\usepackage{graphicx}
\usepackage{amsmath}
\usepackage{amsthm}
\usepackage{booktabs}
\usepackage{algorithm}
\usepackage{algorithmic}
\usepackage{multirow}
\usepackage{array}
\usepackage{colortbl}
\usepackage{ragged2e} 
\usepackage{tabularx}
\usepackage{color,xcolor}
\usepackage{float}
\usepackage{booktabs}
\usepackage{multirow}

\definecolor{cvprblue}{rgb}{0.21,0.49,0.74}
\usepackage[pagebackref,breaklinks,colorlinks,allcolors=cvprblue]{hyperref}

\title{CleanerCLIP: Fine-grained Counterfactual Semantic Augmentation for Backdoor Defense in Contrastive Learning}

\author{Yuan Xun$^{1}$, Siyuan Liang$^{2}$, Xiaojun Jia$^{3}$, Xinwei Liu$^{1}$, Xiaochun Cao$^{4}$ \\
    $^{1}$Institute of Information Engineering, Chinese Academy of Sciences \\
    $^{2}$National University of Singapore \\
    $^{3}$Nanyang Technological University \\
    $^{4}$Sun Yat-sen University-Shenzhen \\
}

\begin{document}
\maketitle
\begin{abstract}
Multimodal contrastive models like CLIP are increasingly vulnerable to data-poisoning backdoor attacks. Existing defense methods primarily target the pretraining phase. However, with the rise of open-source communities, pretrained models are now freely available for download and fine-tuning. These models may carry unknown security risks, posing significant threats to downstream users. This highlights the need for lightweight defense strategies tailored specifically for the fine-tuning stage. Current defenses during fine-tuning include: finetuning with clean data; and using unimodal self-supervised techniques like CleanCLIP, which has represented the state-of-the-art (SOTA). However, these methods rely on strengthening clean feature representations to mitigate attacks, making them ineffective against more stealthy backdoor techniques, such as BadCLIP, which leverage covert toxic features.
To overcome this limitation, we propose a finetuning defense mechanism based on fine-grained counterfactual text semantic augmentation. By modifying small portions of text during fine-tuning, our approach disrupts the association between backdoor triggers and target features.
We evaluate our method against six attack algorithms and conduct comprehensive zero-shot classification on ImageNet1K. Experimental results demonstrate that our method achieves SOTA performance in fine-tuning defense. Specifically, when facing the novel BadCLIP attack, our method surpasses CleanCLIP, reducing the Attack Success Rate (ASR) by 52.02\% in the Top-1 and 63.88\% in the Top-10 classifications.
\end{abstract}    
\section{Introduction}
\label{sec:intro}

 Contrastive learning serves as a powerful learning paradigm aimed at comparing different representations of data, thereby bringing similar samples closer together in the embedding space while pushing dissimilar samples further apart~\cite{chen2020simple,khosla2020supervised,gutmann2010noise}. In addition to its application in single-modal data~\cite{gao2021simcse,chen2022dual,bi2022vision,park2020contrastive}, recent works have extended contrastive learning to multimodal data~\cite{zhang2023generalization,singh2023coarse,yang2022unified}, training on a vast scale of image-text pairs from the web to achieve joint feature representation and matching between images and text. Multimodal contrastive pre-trained models, such as CLIP~\cite{CLIP}, ALIGN~\cite{ALIGN}, and BASIC~\cite{BASIC}, have learned universal representations from large-scale unlabeled data and performed exceptionally well even without task-specific data, as demonstrated by their impressive zero-shot classification performance on ImageNet~\cite{deng2009imagenet}. By fine-tuning these models on specific tasks with a small amount of labeled training samples, high-performance vertical domain applications can be realized quickly. 

\begin{figure*}[t]
    \centering
    \includegraphics[width=0.95\linewidth]{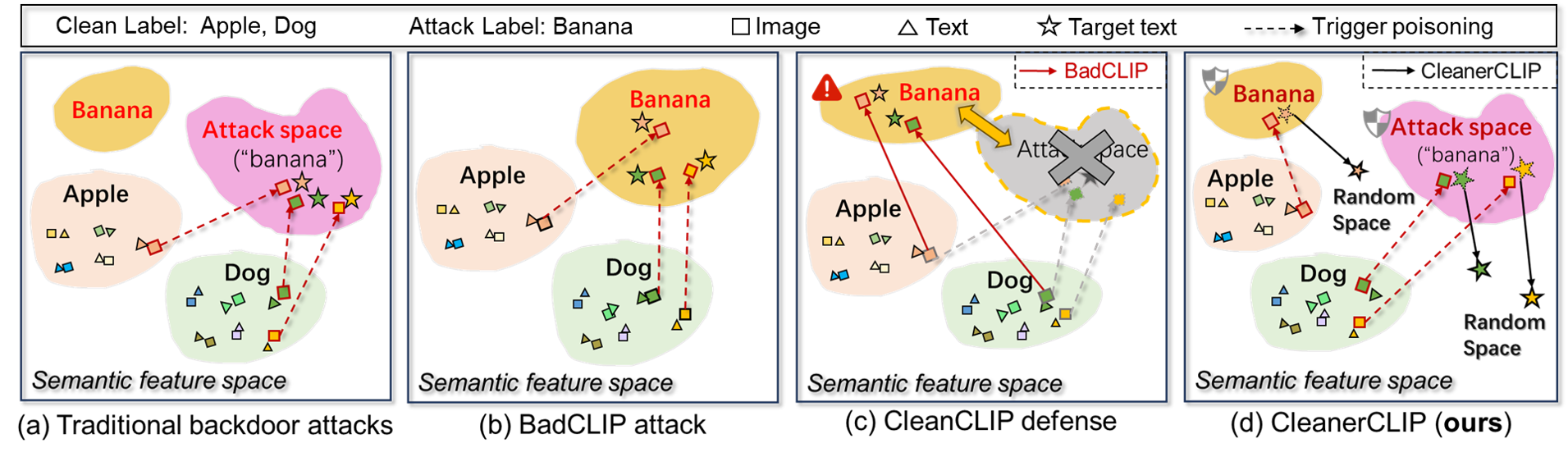}
    \caption{Overview of backdoor attack strategies and defenses. (a) Traditional backdoor attacks form pseudo-semantic clusters by linking visual triggers to specific texts. (b) BadCLIP avoids detection by directly targeting true feature regions without creating pseudo-clusters. (c) CleanCLIP disrupts pseudo-clusters using self-supervised learning. (d) CleanerCLIP enhances defense by generating fine-grained counterfactual subtexts, breaking the semantic link between the trigger and target.}
    \label{fig: motivation}
\end{figure*}
 
 However, recent research has revealed that these models are vulnerable to data-poisoning backdoor attacks~\cite{gao2020backdoor,jia2022badencoder,saha2022backdoor,carlini2021poisoning,li2023embarrassingly}, which can compromise their integrity and reliability. In a backdoor attack, an adversary embeds a trigger into the model, allowing it to misclassify inputs in specific, often harmful ways. This vulnerability poses a serious concern, particularly as these models are increasingly deployed in real-world applications. Existing defense methods primarily target the pretraining phase, aiming to mitigate risks before models are fine-tuned for specific tasks. Notable approaches include CleanCLIP~\cite{bansal2023cleanclip} and RoCLIP~\cite{RoCLIP}, which focus on enhancing the model's robustness against backdoor attacks during this initial stage. However, the rise of open-source communities has facilitated the widespread availability of pre-trained models, many of which may harbor unknown security risks. Users often download and fine-tune these models for personalized applications, inadvertently exposing themselves to potential threats. This scenario underscores the critical need for lightweight defense strategies that can be applied during the fine-tuning stage. 

 Current fine-tuning defenses typically involve two which primarily rely on reinforcing clean feature representations: 1) FT: directly fine-tuning with clean samples. 2) CleanCLIP: employing unimodal self-supervised learning, which can also be adapted for the fine-tuning phase and has achieved SOTA performance. While effective against some known threats, these two approaches become inadequate when facing more covert backdoor techniques, such as BadCLIP~\cite{liang2023badclip}. BadCLIP can exploit hidden toxic features within the clean feature space, evading detection by bypassing the self-supervised defenses of methods like CleanCLIP.

 To clarify the motivation behind our method and its effectiveness, we explore the landscape of backdoor attacks. Traditional backdoor attacks create new feature clusters in the feature space by linking visual triggers to specific texts, thereby assigning new pseudo-semantics to the target text. While these attacks can be effective, they leave distinct traces of pseudo-clusters, making detection and defense more manageable, as shown in Figure~\ref{fig: motivation}(a). CleanCLIP further addresses this vulnerability by incorporating a vision-language self-supervised learning module to disrupt these pseudo-semantic clusters, as shown in Figure~\ref{fig: motivation}(c). While effective against some known threats, CleanCLIP performs inadequately when facing more covert backdoor techniques, such as BadCLIP~\cite{liang2023badclip}, which exploits hidden toxic features to evade detection. Instead of generating new pseudo-clusters, BadCLIP precisely identifies the true feature regions of the target text and adjusts the image trigger to approach these regions, successfully evading the self-supervised enhancements, as shown in Figure~\ref{fig: motivation}(b). This limitation underscores the necessity for a more robust defense mechanism.

To address this novel challenge, we introduce CleanerCLIP, an innovative strategy that utilizes fine-grained counterfactual semantic augmentation to disrupt the potential semantic link between the trigger and the target output, as illustrated in Figure~\ref{fig: motivation}(d). In contrast to CleanCLIP, which primarily focuses on disrupting pseudo-semantic clusters, our approach also generates negative and positive subtexts for a small subset of the clean fine-tuning data. Negative subtexts are created by randomly replacing components of the text's semantics. This random alteration disrupts the semantic binding exploited in potential backdoor attacks, reducing the stability and success rate of the trigger. Meanwhile, positive subtexts preserve the essential semantic features of the original target text, ensuring that the model can accurately process clean data. This dual augmentation process not only lowers the success rate of backdoor attacks but also enhances the overall robustness of the model during fine-tuning.

Our contributions can be summarized as follows:


\begin{itemize} 
\item We identify the limitations of current backdoor defense methods during fine-tuning, particularly their failure to address covert attacks like BadCLIP, highlighting the need for more robust defenses. 
\item We propose CleanerCLIP, a lightweight defense strategy that employs fine-grained counterfactual semantic augmentation to disrupt the trigger-target connection and enhance robustness. 
\item Our method is tested against six attack techniques, and experimental results show that CleanerCLIP significantly reduces attack success rates while maintaining the model's benign accuracy and usability. 
\end{itemize}

\section{Related work and preliminaries}
\label{sec:related_work}
\textbf{Contrastive Language-Image Pre-Training} CLIP~\cite{CLIP}, released by OpenAI, stands as a prominent representative of MCL. Inspired by mapping images and texts into a shared feature embedding space $\mathbb{R}^d$, CLIP enables the model to understand the semantic relationship between them. CLIP involves two encoders: an image encoder $f_I: I \rightarrow \mathbb{R}^d$ and a text encoder $f_T: T \rightarrow \mathbb{R}^d$, which transform the image and text data into representations of dimension $d$. The model is pre-trained through contrastive learning, leveraging vast amounts of internet image-text pairs $\{I_i, T_i\}_{i=1}^{N}$ to learn the associations between images and texts. During training, the CLIP model learns a mapping function that projects images and texts into the same feature space. This is achieved by maximizing the similarity between positive pairs (matching images $I_i$ and texts $T_i$) while minimizing the similarity between negative pairs (mismatched images and texts). This unsupervised joint learning approach enables the CLIP model to achieve superior performance on various visual and language tasks, including image classification, text caption generation, and image retrieval. The mathematical expression for $loss_{Clip}$ can be found in Sec 7 in the supplementary material.

\textbf{Backdoor attacks}
Backdoor attacks generally refer to the implantation of specific trigger patterns during the model training process, which enables the model to perform normally under normal conditions but exhibit abnormal behavior under specific conditions, such as when the input contains images with trigger patterns. In the domain of supervised learning, backdoor attacks have garnered significant attention, with notable works including BadNet~\cite{gu2017identifying}, Blended~\cite{chen2017targeted}, SIG~\cite{liu2020reflection}, WaNet~\cite{nguyen2021wanet}, and SSBA~\cite{li2021invisible}]. Backdoor attacks targeting the CLIP model primarily leverage its capability to learn from multimodal data. Attackers can add image-text pairs containing specific trigger patterns to the training data, allowing the model to learn the association between these trigger patterns and abnormal behaviors.  Within the domain of MCL,~\cite{carlini2021poisoning} pioneered the revelation of its vulnerability to backdoor attacks, demonstrating a successful attack on CLIP, for instance, by poisoning merely $0.01 \%$ of the data. Concurrently,~\cite{yang2023data} delved into the impact of attacks from different modalities on MCL. Additionally, research on attacks against self-supervised learning (SSL), a broader category, is also ongoing, exemplified by BadEncoder~\cite{jia2022badencoder}, GhostEncoder~\cite{wang2024ghostencoder}, and distribution-preserving attacks~\cite{tao2023distribution}. The details about data-poisoning backdoor attacks on CLIP are shown in Sec 6 in the supplementary material.

\textbf{Backdoor Defenses on CLIP}
To address these threats mentioned above, some researchers have borrowed backdoor defense techniques from supervised learning~\cite{zhu2023enhancing,zhu2024neural} to mitigate the backdoor effects in MCL models. Currently, defense techniques for MCL can be categorized into two groups based on whether the defender can access the poisoned dataset: \ding{172} defenders can access the entire poisoned dataset~\cite{yang2023better,yang2024robust,bansal2023cleanclip}; \ding{173} defenders can only access the poisoned model~\cite{bansal2023cleanclip}. The former approach, which allows for complete retraining of large models with various data augmentation strategies, can achieve strong defense performance, such as RoCLIP~\cite{yang2024robust}. However, in reality, the feasibility of attackers manipulating the training set is low, as they cannot guarantee that their carefully crafted poisoned data will be incorporated into large-scale training sets. Therefore, a more realistic attack strategy is to perform low-cost fine-tuning of existing pre-trained large models with dirty data. As a result, defense techniques targeting the fine-tuning phase are necessary, which is the attack-defense scenario addressed in this paper. A representative example of such defenses is CleanCLIP~\cite{bansal2023cleanclip}. Specifically, CleanCLIP introduces a self-supervised loss based on multimodal data augmentation, which fine-tunes a clean dataset to reduce the impact of backdoor models. Their self-supervised loss $loss_{SS}$ and total fine-tuning loss $loss_{CClip}$ can be found in Sec 7 in the supplementary material.

\section{Methodology}
\label{sec: method}
\subsection{Threat model}
\begin{figure*}[t]
    \centering
    \setlength{\belowcaptionskip}{-0.5cm} 
    \includegraphics[width=0.93\linewidth]{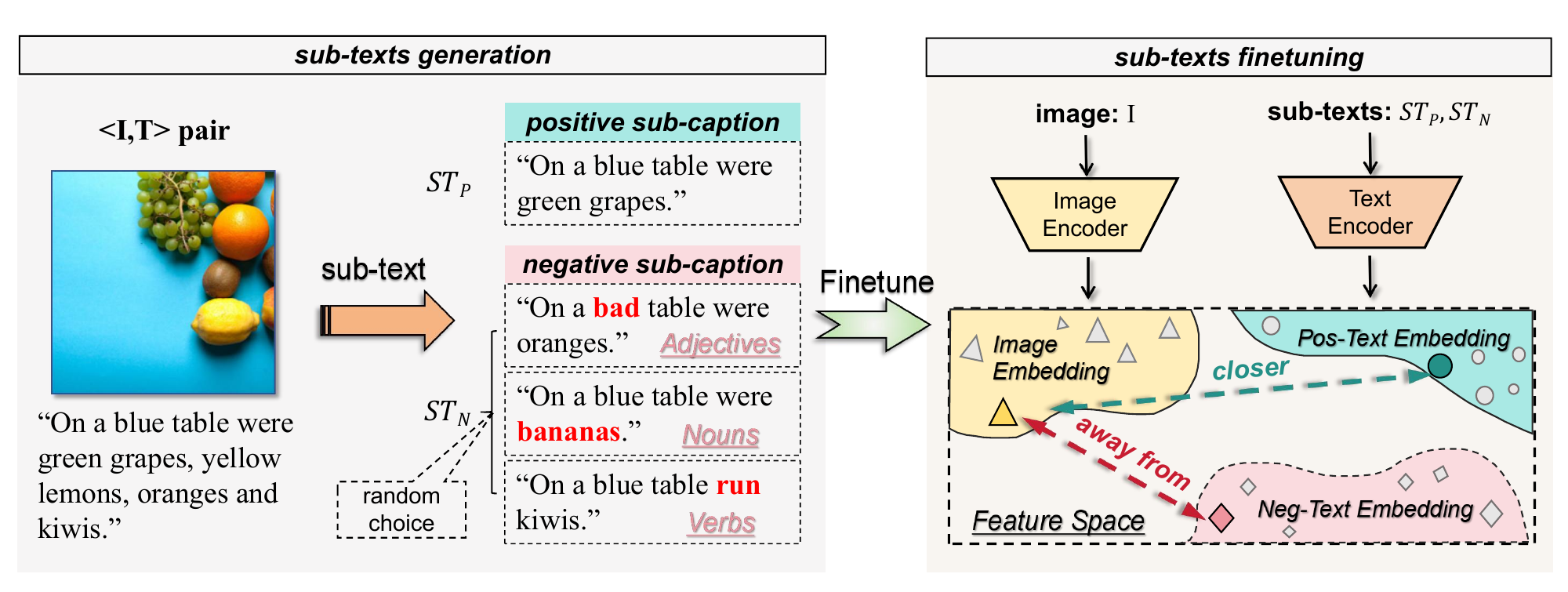}
    \caption{The framework of our CleanerCLIP, illustrating the process of \textbf{factual}(positive) and \textbf{counterfactual}(negative) sub-text generation and fine-tuning. For each raw caption, one of three counterfactual generation strategies is randomly applied. Text augmentation is selectively performed on a small portion of samples during each fine-tuning epoch to ensure minimal computational overhead. }
    \label{fig:TA_framework}
\end{figure*}
\textbf{Adversary Objective}: By polluting the original dataset, the model can generate malicious adversarial text specified by the adversary for any input image embedded with a trigger. During zero-shot testing, the attack objective manifests as poisoned images will be misclassified as the adversarial category, while other benign images will be correctly classified.

\textbf{Adversary Capability}: We assume the attacker possesses knowledge of the model's structure, training algorithm, and the hyper-parameters used by the victim, but they can't directly modify the training process. While the attacker lacks access to the entire dataset, they can inject a small number of poisoned samples into the training dataset. Furthermore, the attacker can poison pre-trained MCL models by fine-tuning with carefully crafted dirty datasets and distributing them through various channels on the internet, thereby creating uncontrollable risks for downstream tasks.

\textbf{Defender Capability}: The defender has access to the model structure and a clean fine-tuning dataset, and can retrain the model to mitigate the attack. However, the defender cannot access the model’s parameters 

\subsection{Counterfactual Text Augmentation}
To address the inadequacies of existing defenses like CleanCLIP against covert backdoor attacks, particularly those employing stealthy trigger features, we propose CleanerCLIP, a fine-grained counterfactual text augmentation strategy. Our approach recognizes the necessity of both preserving clean sample characteristics and disrupting the malicious semantic links exploited by potential backdoor attacks. As illustrated in the left part of Figure~\ref{fig:TA_framework}, our method consists of two main parts: (1) Factual positive sub-caption generation,  which ensures the integrity of the original clean data; (2) Counterfactual negative sub-caption generation, which actively undermines the stealthy backdoor triggers. For convenience, we will refer to positive sub-captions as factual sub-captions and negative sub-captions as counterfactual sub-captions in the following descriptions. Assuming the image-text dataset used by CLIP finetuning is $\mathcal{D}_{ft}$, we annotate each sample as $( I_i, T_i ) \in \mathcal{D}_{ft}$, where $I_i$ is the image and $T_i$ is its associated caption. And we generate $[ST^i_p, ST^i_n]$ for each $T_i$, representing the positive sub-caption and negative sub-caption respectively.

\textbf{Factual: Positive sub-caption generation}
For each text sample $T_i$, we decompose it by first identifying its core semantic components, focusing on relational verbs and key nouns. To generate the positive sub-caption $ST^i_p$, we retain the primary relational verb and key nouns while selectively simplifying or omitting adjectives, ensuring that the core semantic meaning of the original text remains intact. Using SceneGraphParser~\footnote{\url{https://github.com/vacancy/SceneGraphParser}}, we organize the positive sub-caption in the following template, preserving the correct relationships between subjects and objects:

\textit{< ( Adjective of the subject ) + Subject + Relational Verb + ( Adjective of the object ) + Object >}

When the original text is too simple or contains minimal content (e.g., “a picture of an apple”), we preserve only the most critical word, such as "apple", as the positive sub-caption to retain the core meaning.

\textbf{Counterfactual: Negative sub-caption generation}
For each sub-caption, we perform random semantic replacement operations aimed at disrupting the binding between the trigger and the target text. Since we cannot precisely know which entity, attribute, or relationship the adversary might exploit, our replacements are comprehensive and cover all possible elements. We apply three types of replacement operations, and one is selected randomly for each sample: \ding{172} Replace the adjectives associated with the subject and object. If no adjectives are present, this step is skipped. \ding{173} Replace the relational verbs. If missing, another replacement method is chosen. \ding{174} Replace the subject and object nouns.

Considering the large, noisy, and uncurated nature of pre-trained models’ training data, which captures a rich and diverse data distribution, we augment text using a combination of WordNet~\cite{fellbaum2010wordnet} and the large language model ChatGPT~\cite{chatGPT}. The latter helps generate a diverse repository of alternative words, ensuring that negative subtexts are loosely distributed in feature space. Each replacement repository contains 3000 terms, ranging from common to rare words, ensuring maximum disruption of the backdoor trigger's semantic binding.

\subsection{CleanerCLIP: Fine-grained Counterfactual Semantic Finetuning}
\begin{algorithm}[t]
    \caption{CleanerCLIP Finetuning Algorithm}
    \begin{algorithmic}[1]
    \REQUIRE The benign finetune image-text pairs $\{I_i, T_i\} \in \mathcal{D}_{ft}$, the fine-tuning batch size $N$, the image encoder $f_I$ , the text encoder $f_T$, the number of texts need to be augmented $K$, the generation function of positive and negative sub-texts $G_p(\cdot)$ and $G_n(\cdot)$, the weight of two loss functions $\alpha$ and $\beta$.
    \FOR{epoch from 1 to E}
        \STATE Random select $K$ image-text pairs from $\mathcal{D}_{ft}$, and generate associated positive and negative sub-captions: $ST^i_p = G_p(T_i), ST^i_n=G_n(T_i)$
        \STATE Get feature embeddings of $\{I_i, T_i, ST^i_p, ST^i_n\}$: \\
        $z^I_i = f_I(I_i)$, $z^T_i = f_T(T_i)$, $z^{p}_i = f_T(ST^i_p)$, $z^{n}_i = f_T(ST^i_n)$.
        \STATE $Loss = loss_{Cleaner} = \alpha \cdot loss_{CClip} + \beta \cdot loss_{p-n}$
        
    \ENDFOR
    \end{algorithmic}
    \label{algo:TA_Cleaner}
\end{algorithm}

Previous defense strategies have largely concentrated on leveraging image and text self-supervised learning to counter backdoor triggers. However, our analysis reveals a critical limitation in CleanCLIP: its self-supervision strength is insufficient to defend against meticulously crafted triggers that exploit existing clean features without introducing additional toxic embedding clusters. Specifically, while CleanCLIP reinforces clean feature representations, it fails to address subtle manipulations of these features by attackers who strategically leverage the clean features themselves. This gap highlights the need for a more robust defense against advanced, stealthy attack techniques during finetuning.
To overcome this limitation, we propose a fine-grained text semantic augmentation approach that incorporates both positive and negative sub-captions during the fine-tuning process. This method optimizes text features by employing alternating optimization between self-supervised learning and adversarial sample generation. By reinforcing the robustness of text representations without compromising the expressiveness of clean samples, this approach provides enhanced protection against image-based backdoor triggers, aligning with the lightweight defense mechanism we advocate for.

We avoid fine-grained semantic augmentation on the whole dataset during fine-tuning to prevent disrupting the alignment of clean images and texts, which may lower the downstream zero-shot accuracy. We randomly select $K$ samples from all text data for fine-grained augmentation, obtaining $K$ augmented data, denoted as $\{I_i,T_i,ST^i_p,ST^i_n\}^{K}_{i=1}$. This random selection approach maximally retains the original feature expression capability's generalization on downstream tasks while achieving our defense objectives, and effectively minimizing the additional computational cost of the fine-tuning defense. The mapping of these $K$ data points in the feature space is denoted as $\{z^I_i,z^T_i,z^{p}_i,z^{n}_i\}^{K}_{i=1}$. Note that here, $z^p_i$ and $z^n_i$ represent the feature embeddings of positive and negative subtexts, respectively.

Our factual-counterfactual finetuning loss function consists of two parts: the $loss_{i2t}$ measures the similarity between positive sample images and text and the dissimilarity between negative sample texts and images, thereby minimizing the information difference between positive sample images and text. The $loss_{t2i}$ measures the similarity between positive sample text and images and the dissimilarity between negative sample images and text, thereby minimizing the information difference between positive sample text and images. Both parts jointly optimize the consistency of multi-modal embedding space. The specific mathematical expressions are as follows:



\begin{equation}
  \resizebox{\columnwidth}{!}{$
    loss_{i2t} = -\frac{1}{K} \sum_{i=1}^{K} \log \left( \frac{\exp\left( \langle z^I_i, z^{p}_i \rangle / {t_p} \right)}{\sum_{j=1}^{K} \exp\left( \langle z^I_i , z^{p}_j \rangle / {t_p}\right) + \sum_{k=1}^{K} \exp\left( \langle z^I_i , z^{n}_i \rangle / {t_n}\right)} \right)
  $}
  \label{Li2t}
\end{equation}
\begin{equation}
  \resizebox{\columnwidth}{!}{$
    loss_{t2i} = -\frac{1}{K} \sum_{i=1}^{K} \log \left( \frac{\exp\left( \langle z^{p}_i , z^I_i \rangle / {t_p}\right)}{\sum_{j=1}^{K} \exp\left( \langle z^{p}_j , z^I_i \rangle / {t_p}\right) + \sum_{k=1}^{K} \exp\left( \langle z^{n}_i , z^I_i \rangle / {t_n}\right)} \right)
  $}
  \label{Lt2i}
\end{equation}
\begin{equation}\label{L_ta}
  loss_{p-n} = ( loss_{i2t} + loss_{t2i} ) / 2 .
\end{equation}
Here, $t_p$ and $t_n$ are the temperature parameters for positive and negative samples, which control the sensitivity of the loss function to positive and negative samples by adjusting the weight of the similarity score. Specifically, increasing $t_p$ enhances the sensitivity of the similarity score of positive samples, leading the loss function to focus more on the differences between positive samples, which may result in less ideal defense effects. Similarly, increasing $t_n$ enhances the sensitivity of the similarity score of negative samples, which may lead to excessive learning of negative samples by the model, ignoring the similarity between positive samples and reducing the model's generalization ability. Therefore, the setting of these two hyper-parameters $t_p$ and $t_n$ also has a certain degree of influence on the adversarial learning between positive and negative samples. Hence, our total loss function $loss_{Cleaner}$ can be described as follows:
\begin{equation}\label{L_TA}
    loss_{Cleaner} = \alpha \cdot loss_{CClip} + \beta \cdot loss_{p-n},
\end{equation}
where $\alpha$ and $\beta$ are hyper-parameters, representing the weight of $loss_{CClip}$ and $loss_{p-n}$ respectively. Finally, our complete finetuning steps are given in Algorithm~\ref{algo:TA_Cleaner}.



\section{Experiments}
\label{experiments}
\subsection{Setup}
\textbf{Dataset and models} 
As a defense technique during the fine-tuning phase, we adopted the fine-tuning setting of~\cite{bansal2023cleanclip}. We utilized the open-source CLIP model from OpenAI~\cite{radford2021learning} as the pre-trained clean model, which is trained on a dataset containing 400 million image-text pairs. We selected 500,000 image-text pairs (CC500K) as our fine-tuning dataset from the CC3M dataset~\cite{CC3M}. Following~\cite {bansal2023cleanclip}, we use the ResNet-50 model as the CLIP vision encoder and a transformer as the text encoder during fine-tuning. We conducted our experiments using an A100 GPU.
\begin{table*}[t]
\renewcommand\arraystretch{1.5}
\centering
\resizebox{0.95\textwidth}{!}{%
\begin{tabular}{cccccccccccccc}
\hline
\multicolumn{2}{c}{\multirow{2}{*}{Methods}} & \multicolumn{2}{c}{BadNet} & \multicolumn{2}{c}{Blended} & \multicolumn{2}{c}{SIG} & \multicolumn{2}{c}{WaNet} & \multicolumn{2}{c}{SSBA} & \multicolumn{2}{c}{BadCLIP} \\
\cmidrule(r){3-4} \cmidrule(r){5-6} \cmidrule(r){7-8} \cmidrule(r){9-10} \cmidrule(r){11-12} \cmidrule(r){13-14}
\multicolumn{2}{c}{}                         & BA $\uparrow$        & ASR $\downarrow$           & BA $\uparrow$      & ASR $\downarrow$             & BA $\uparrow$    & ASR $\downarrow$          & BA $\uparrow$      & ASR $\downarrow$          & BA $\uparrow$  & ASR $\downarrow$        & BA $\uparrow$         & ASR $\downarrow$     \\ \hline

\multicolumn{1}{c|}{\multirow{4}{*}{Top-1}}  & NoDefense & 59.32    & 91.28            & 59.35    & 68.7              & 59.59  & 80.08           & 59.53    & 91.83             &  58.48             &  50.08   & 58.77     & 99.27               \\
\multicolumn{1}{c|}{}                        & FT   & 54.98    &  62.14    & 54.63    & 40.45          & 52.69  &7.71   &  52.72  &  0.57         & 55.73              & 3.82     &  54.05    &96.22                 \\ 
\multicolumn{1}{c|}{}                        & CleanCLIP   & 52.62    & 1.88             & 52.68    & 13.29             & 52.17  & 10.06           & 52.74    & 0.53           &   55.02             &  3.87    & 52.61     & 69.87                \\ 
\multicolumn{1}{c|}{}                        & \textbf{CleanerCLIP}        & 52.64    & \textbf{0.45}    & 52.61    & \textbf{12.84}    & 52.36  & \textbf{1.41}   & 52.61    & \textbf{0.15}     &  54.91    &  \textbf{1.06} & 51.29     & \textbf{17.85}         \\ \cmidrule{1-1} \cmidrule{2-14}
\multicolumn{1}{c|}{\multirow{4}{*}{Top-3}}  & NoDefense & 79.8     & 97.16            & 80.02    & 81.09             & 79.98  & 90.12           & 80.05    & 96.77           &  78.96       & 77.12   & 79.69     & 99.66                   \\ 
\multicolumn{1}{c|}{}                        & FT   &  76.20   &81.64             & 76.92    &  58.94         &74.65   & 18.52           & 74.51   &  1.62         & 76.92           &13.67      & 76.70     & 98.54                \\ 
\multicolumn{1}{c|}{}                        & CleanCLIP   & 74.35    & 5.82             & 75.48    & 26.44             & 74.42  & 22.48           & 74.09    & 1.58          & 76.47           &  12.81    & 74.71     & 82.72                 \\  
\multicolumn{1}{c|}{}                        & \textbf{CleanerCLIP}        & 73.76    & \textbf{1.61}    & 75.54    & \textbf{24.37}    & 73.67  & \textbf{4.86}   & 73.54    & \textbf{0.49}   & 76.16     &  \textbf{5.94}  & 74.37     & \textbf{20.19}         \\ \cmidrule{1-1} \cmidrule{2-14}
\multicolumn{1}{c|}{\multirow{4}{*}{Top-5}}  & NoDefense & 86.19    & 98.39            & 86.14    & 85.53             & 86.3   & 93.1            & 86.16    & 98.09             &  85.51       &  85.01     & 85.94     & 99.74            \\ 
\multicolumn{1}{c|}{}                        & FT   & 83.87    & 88.02            &83.61     & 66.94          &81.98   &  26.46          &81.82    & 2.78          &   84.76          &22.13      &  83.44    & 98.97                 \\ 
\multicolumn{1}{c|}{}                        & CleanCLIP   & 81.73    & 9.45             & 81.69    & 34.79             & 81.98  & 30.66           & 81.97    & 2.10         & 83.58         & 20.16  & 81.96     & 87.38            \\ 
\multicolumn{1}{c|}{}                        & \textbf{CleanerCLIP}      & 80.98    & \textbf{2.97}    & 81.71    & \textbf{33.88}    & 80.96  & \textbf{7.94}   & 81.73    & \textbf{0.96} & 84.67   & \textbf{8.83}    & 81.43     & \textbf{24.53}            \\ \cmidrule{1-1} \cmidrule{2-14}
\multicolumn{1}{c|}{\multirow{4}{*}{Top-10}} & NoDefense & 92.08    & 99.19            & 92.23    & 90.14             & 92.13  & 96.11           & 92.11    & 99.12               & 91.44     &92.16     & 91.99     & 99.83             \\
\multicolumn{1}{c|}{}                        & FT   & 89.99    &   94.01          &89.95    &76.74           & 89.22  &39.98            &88.97    &  5.75         & 90.97         & 37.14     &   89.93   & 99.37                 \\ 
\multicolumn{1}{c|}{}                        & CleanCLIP   & 88.92    & 17.11            & 88.99    & 47.97             & 89.11  & 44.89           & 89.11    & 4.58          & 90.21     &  33.29    & 88.99     & 91.96        \\ 
\multicolumn{1}{c|}{}                        & \textbf{CleanerCLIP}       & 88.91    & \textbf{6.77}    & 88.86    & \textbf{46.72}    & 89.13  & \textbf{15.70}  & 89.12    & \textbf{2.19}    & 90.02   & \textbf{10.27}     & 88.81     & \textbf{28.08}         \\ \cmidrule{1-1} \cmidrule{2-14}
\end{tabular}%
}
\caption{The defense performance of Top-k BA (\%) and ASR (\%), targeting multi backdoor attacks.}
\label{tab:TAresults}
\vspace{-0.4cm}
\end{table*}

\textbf{The victim models generation} 
We utilized the CC500K dataset to simulate the adversary's attack process. Specifically, we randomly selected 1,500 samples from CC500K for various types of backdoor attacks, embedding triggers into the images. The corresponding text was modified to target specific categories using a predefined template, while the remaining samples were kept unchanged. This contaminated dataset was then used to fine-tune the pre-trained CLIP model. For fine-tuning, we employed a batch size of 128, an iteration count of 5, and a base learning rate of \(1 \times 10^{-6}\). The learning rate was warmed up over 10,000 steps, using AdamW as our optimizer with a weight decay of 0.1. The Adam momentum factor and RMSProp factor were set to 0.9 and 0.999, respectively, with an epsilon value of \(1 \times 10^{-8}\). And the attack target label is ``banana".

\textbf{Defense finetuning} 
We employed the CC500K dataset to conduct clean fine-tuning (FT), CleanCLIP, and our proposed CleanerCLIP. For both methods, we utilized a batch size of 64, an iteration count of 10, and AdamW as the optimizer. The learning rate was warmed up over 10,000 steps, with a weight decay of 0.1 for the optimizer. The Adam momentum factor and RMSProp factor were set to 0.9 and 0.999, respectively, with an epsilon value of \(1 \times 10^{-8}\). The base learning rate for both methods was set to \(4.5 \times 10^{-6}\).

\textbf{Evaluation metrics} 
Following~\cite{RoCLIP,bansal2023cleanclip} and most attacks like~\cite{liang2023badclip}, we adopt benign accuracy (BA, $\uparrow$) and attack success rate (ASR, $\downarrow$) as our evaluation metrics. For BA, a higher value indicates superior clean performance, while for ASR, a lower value reflects better defense performance. These metrics are used to assess defense strategies across two common tasks: zero-shot classification on the ImageNet-1K validation set and linear probing. In the linear probing test, the feature extraction layers remain fixed, and only the linear layer is trained on clean images from the training set, followed by testing on the validation set.

To comprehensively assess the impact of our defense on CLIP performance, we use \textbf{Top-k} ($k=1,3,5,10$) evaluations for both BA and ASR. Top-k accuracy considers not only the highest probability class predicted by the model but also other high-probability classes, thus offering a more accurate reflection of the model's generalization capacity in multi-class tasks. This consideration has been widely used in prior works on the CLIP community, such as EVA-CLIP~\cite{sun2023evaclip}, CosmoCLIP~\cite{imam2024cosmoclip}, and CEIA~\cite{xu2024ceia}. However, to our knowledge, previous backdoor defense works targeting CLIP have not reported Top-k metrics, leading to incomplete performance evaluation. Therefore, by incorporating Top-k BA/ASR evaluations, our work not only provides a more thorough investigation into CLIP’s backdoor vulnerabilities but also demonstrates state-of-the-art performance by significantly reducing Top-k ASR.
\begin{table*}[t]
\centering
\resizebox{0.9\textwidth}{!}{%
\begin{tabular}{cccccccccc}
\toprule
Datasets     & CIFAR10 & CIFAR100 & ImageNet1K & DTD  & STL10 & SVHN  & Food101 & OxfordIIITPet  & RenderedSST2 \\
\midrule
FT  &  \textbf{83.49}  & 58.08   &  60.58   &   65.27        & 95.10      & 48.78     &  82.22       &\textbf{77.77}        &  70.27         \\
CleanCLIP    & 83.07   & \textbf{60.33}    & \textbf{72.46}      & 65.74               &95.59       &50.54       & 82.04        &  76.70                          &  \textbf{70.35}            \\
CleanerCLIP         & 83.32  & 60.01    & 72.37      & \textbf{65.78}               & \textbf{95.83}      & \textbf{51.06}      & \textbf{83.01}        &  77.21                          & \textbf{70.35}\\
\bottomrule
\end{tabular}%
}
\caption{The linear-probe classification accuracy (\%) on a series of datasets.}
\label{tab:linear_probe}
\end{table*}

\subsection{CleanerCLIP performance}
Similar to~\cite{bansal2023cleanclip} and~\cite{RoCLIP} of pretraining defense methods, we conduct zero-shot testing on ImageNet1K to evaluate our performance. We utilize six attack methods to generate victim models: BadNet~\cite{gu2017identifying}, Blended~\cite{chen2017targeted}, SIG~\cite{liu2020reflection}, WaNet~\cite{nguyen2021wanet}, SSBA~\cite{li2021invisible}, and BadCLIP~\cite{liang2023badclip}. Among them, the first five are classic backdoor attack methods in supervised learning, while BadCLIP is a recently developed attack technique specifically tailored for CLIP. For each attack method, we randomly select 1500 images from CC500K for poisoning and subsequently finetune to generate poisoned models. We apply FT, CleanCLIP, and CleanerCLIP defenses separately to these six poisoned models and obtain the Top-k (k=1,3,5,10) BA (\%) and ASR (\%) after defense finetuning. Our final results are presented in Table~\ref{tab:TAresults}. In our implementation of CleanerCLIP, for the first five attack methods, we randomly selected 1,000 images per iteration for positive and negative subtext generation and finetuning. However, for BadCLIP, we randomly sampled 3,000 images for defense, as this is an exceptionally potent attack method where a smaller sample size would be insufficient to generate a defense boundary to resist the proximity of poisoned image features.

\textbf{Regarding Top-1 performance} 
Compared to the victim model (NoDefense), fine-tuning (FT) improves defense, but certain vulnerabilities remain. For instance, the Top-1 ASR for the BadNet attack is still high at 62.14\%, reflecting limitations in effectively countering attacks. CleanCLIP shows stronger defensive capabilities, but our CleanerCLIP reduces ASR even further, achieving near-zero Top-1 ASR against both BadNet and WaNet (from 91.28\% to 0.45\% and from 91.83\% to 0.15\%, respectively). Against BadCLIP, which poses significant challenges for CleanCLIP, our method also markedly weakens its effectiveness. BadCLIP’s optimization aligns poisoned images closely with target text features, creating a strong trigger association. CleanCLIP’s EDA-based augmentation does not alter semantic content sufficiently to disrupt this binding, allowing triggers to persist within clean features. Our CleanerCLIP, however, uses random counterfactual semantic enhancement to disrupt trigger connections, while factual positive subtexts strengthen defense by preserving clean sample features during fine-tuning.


\textbf{Regarding Top-k performance} 
In zero-shot classification, Top-1 accuracy reflects the model's success in identifying the correct class as the top prediction, while Top-k accuracy assesses whether the true class is among the top-k predictions, thus offering a broader evaluation. Given the inherent challenges in zero-shot tasks, \textbf{Top-k accuracy often provides a more comprehensive measure of performance}, as it takes into account the model's predictive capability across multiple potentially correct classes. We report BA and ASR for Top-3, Top-5, and Top-10 in Table~\ref{tab:TAresults}. Unlike NoDefense, FT, and CleanCLIP, our CleanerCLIP avoids overfitting and shows improved defense even across broader prediction scopes. Notably, when FT and CleanCLIP almost fail against BadCLIP (Top-10 ASR at 99.37\% and 91.96\%, respectively), our method significantly lowers ASR by 71.75\% relative to the original victim model, underscoring the robustness of our approach.

\textbf{About BA}
Due to our defense being a finetuning-based strategy and the limitation of relevant computational resources, fine-tuning a large pre-trained model with a small dataset (3 million VS. 500K) will inevitably affect the capability of clean feature alignment, i.e., the BA performance. Nevertheless, it is noteworthy that, compared with CleanCLIP, a similarly fine-tuning approach, our proposed CleanerCLIP significantly reduces the ASRs while maintaining almost the same BAs as theirs, as shown in Table~\ref{tab:TAresults}.

\textbf{The availability of CleanerCLIP}
We evaluated CleanerCLIP using linear-probe methods on a series of datasets introduced by~\cite{kornblith2019better} to investigate whether it negatively impacts the model's usability and transfer performance. For this evaluation, we tested models subjected to BadNet poisoning and defenses, with a learning rate of $1e-3$ during linear probe training. The corresponding test results are shown in Table~\ref{tab:linear_probe}. As observed, we achieved test results comparable to CleanCLIP, indicating that we significantly reduced the ASR without compromising the model's performance and transferability. More details about the dataset information are shown in Sec 8 in the supplementary material.
\begin{figure*}[ht]
	\centering  
	\begin{subfigure}{0.32\linewidth}  
		\includegraphics[width=\linewidth]{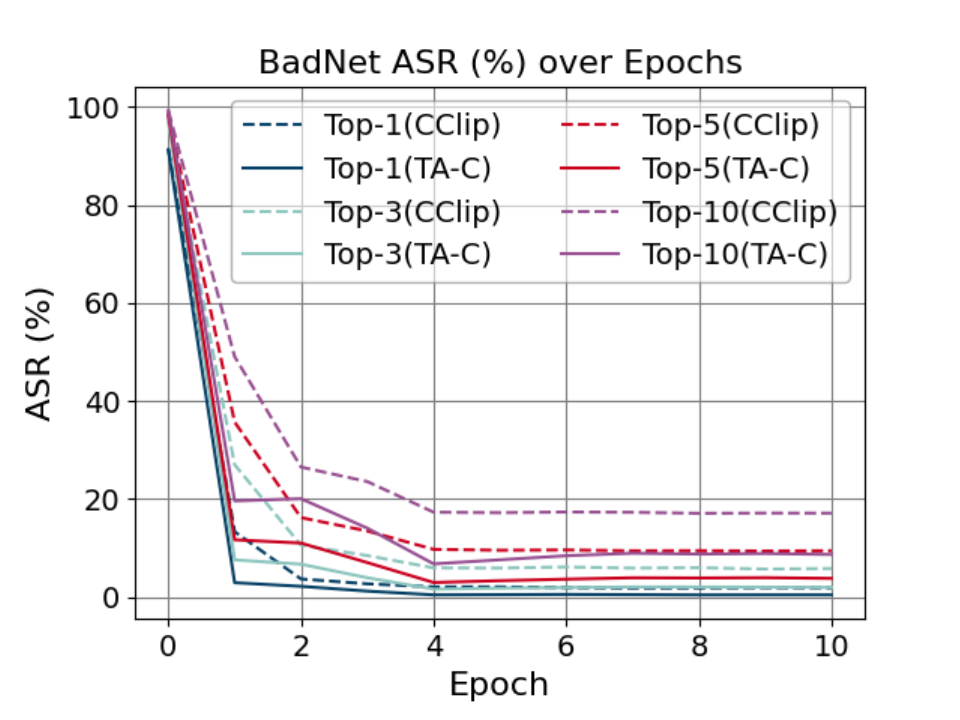} 
		\caption{BadNet} 
		\label{fig:asr_badnet}
	\end{subfigure}
	\hfill  
	\begin{subfigure}{0.32\linewidth}  
		\includegraphics[width=\linewidth]{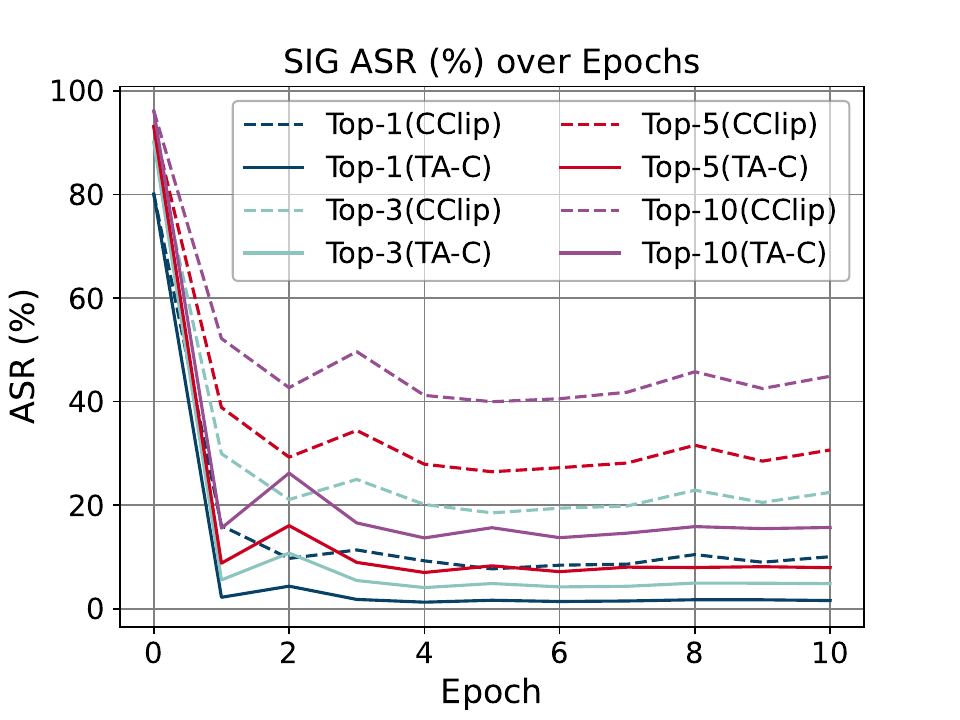} 
		\caption{SIG}
		\label{fig:asr_sig}
	\end{subfigure}
	\hfill  
	\begin{subfigure}{0.32\linewidth}  
		\includegraphics[width=\linewidth]{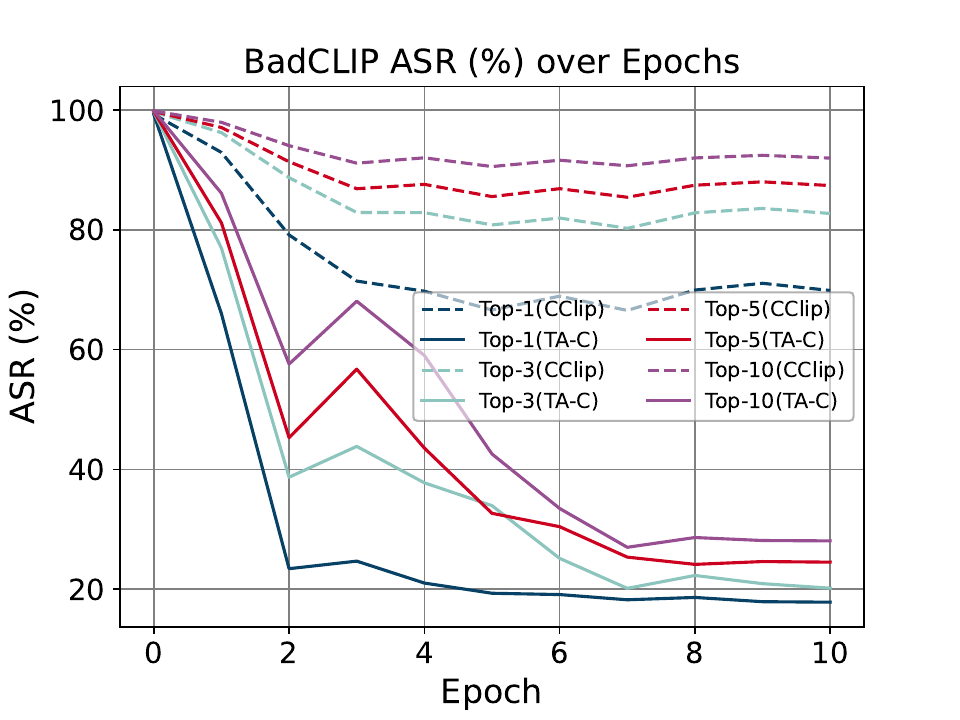} 
		\caption{BadCLIP} 
		\label{fig:asr_badclip}
	\end{subfigure}
	\caption{The decline curve of Top-k ASR (\%) over epochs of CleanCLIP and our CleanerCLIP, on different backdoor attacks.}
	\label{fig:asr_plt}
	\vspace{-0.5cm}
\end{figure*}
\begin{figure*}[ht]
	\centering  
	\begin{subfigure}{0.32\linewidth}  
		\includegraphics[width=\linewidth]{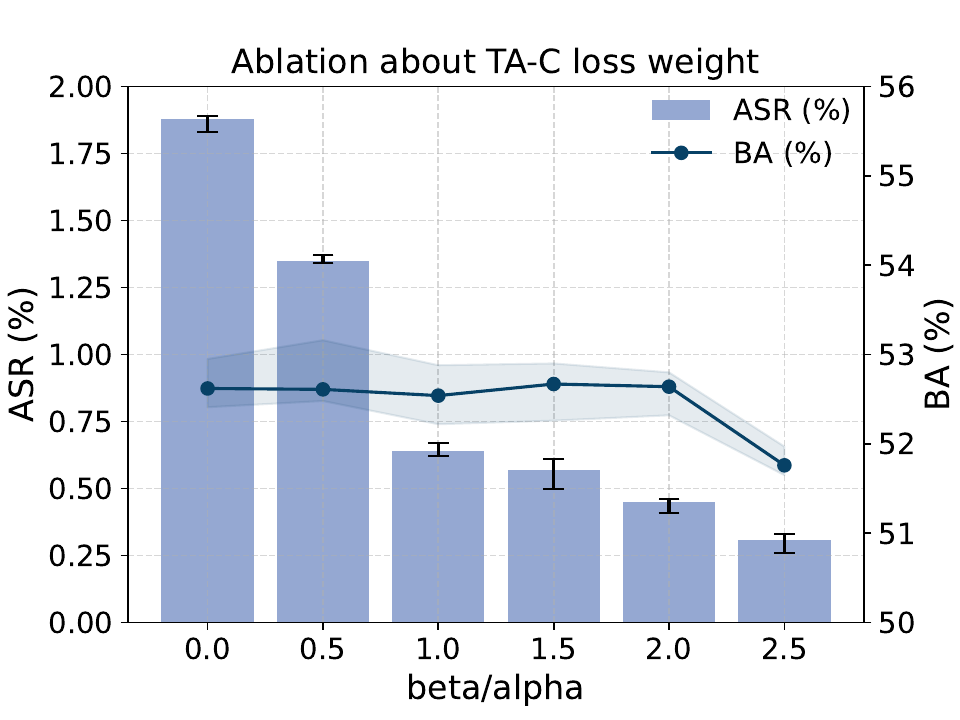}
		\caption{The $loss_{p-n}$ weight} 
		\label{fig:loss_weight}
	\end{subfigure}
	\hfill  
	\begin{subfigure}{0.32\linewidth}  
		\includegraphics[width=\linewidth]{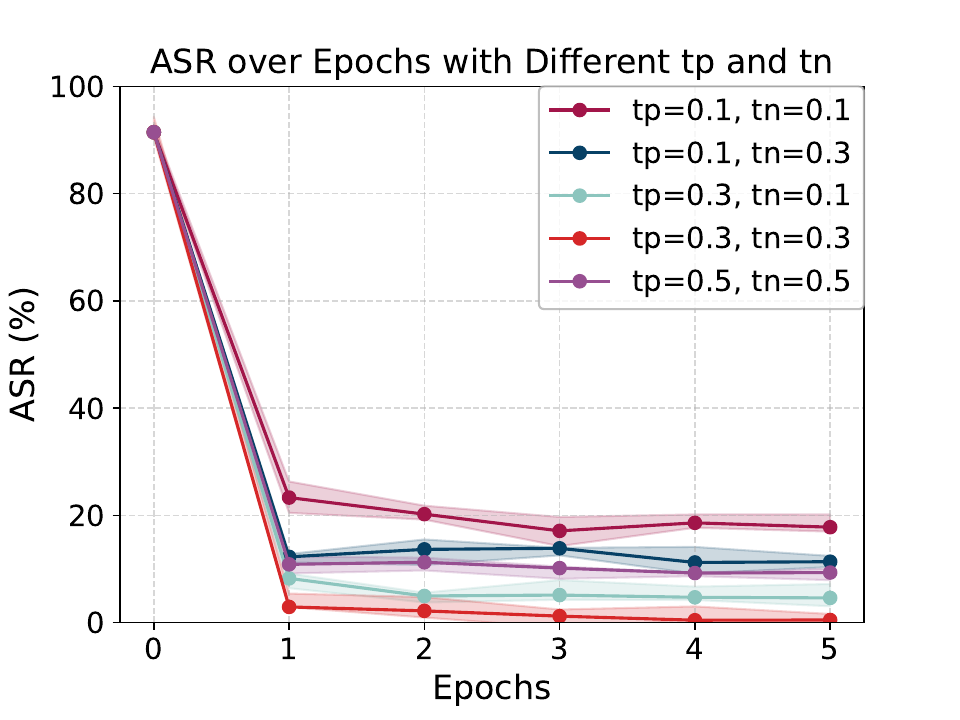} 
		\caption{Temperature factors} 
		\label{fig:temp_factors}
	\end{subfigure}
	\hfill  
	\begin{subfigure}{0.32\linewidth}  
		\includegraphics[width=\linewidth]{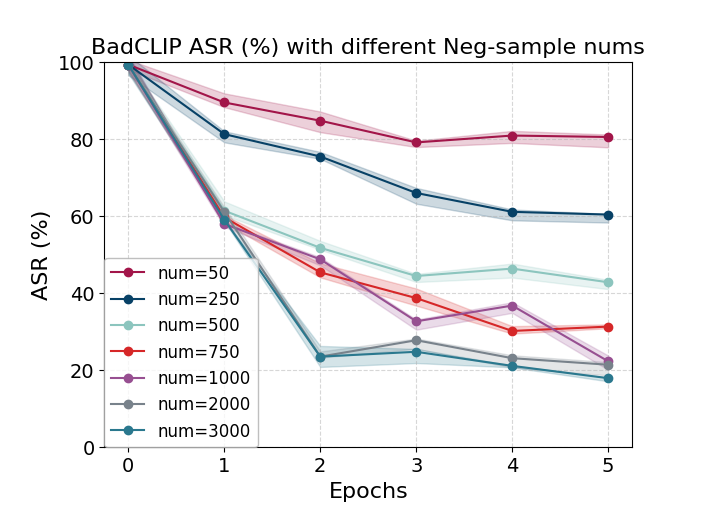} 
		\caption{Neg-sample nums} 
		\label{fig:TA_num}
	\end{subfigure}
	\caption{(a) The Top-1 ASR (\%) and BA (\%) with different $loss_{p-n}$ weight $\beta / \alpha$. (b) The Top-1 ASR (\%) over epochs with different pos/neg temperature targeting BadNet attack. (c) The Top-1 ASR (\%) of BadCLIP over epochs with different Neg-sample numbers (the number of texts we apply CleanerCLIP in each epoch).}
	\label{fig:TA-num}
	\vspace{-0.4cm}
\end{figure*}

\textbf{The ASR during defense epochs}
Beyond the ultimate defense consequence, our CleanerCLIP achieves faster and better defense performance compared to CleanCLIP, as illustrated in Figure~\ref{fig:asr_plt}. We present in Figure~\ref{fig:asr_plt} the ASR trends of CleanCLIP (CClip) and CleanerCLIP with increasing epochs for BadNet, SIG, and BadCLIP attacks. It is observable that CleanerCLIP significantly reduces ASR often within the first epoch and converges relatively steadily thereafter. This implies that merely with the cost of fine-tuning a few thousand additional samples, CleanerCLIP can achieve faster and superior defense performance compared to CleanCLIP.

\subsection{Analysis}
\textbf{Ablation} As shown in Table~\ref{tab:TAresults}, Table~\ref{tab:linear_probe} and Figure~\ref{fig:asr_plt}, when we introduced fine-grained augmentation of positive and negative sub-samples, indicated by the addition of \( loss_{p-n} \), our CleanerCLIP significantly improved defense performance compared to the original CleanCLIP, without compromising the model's performance on clean samples.

\textbf{The \(loss_{p-n}\) weight} 
We evaluated the impact of the proposed \( loss_{p-n} \) on the overall loss for defense performance, as shown in Eq.~\ref{L_TA}. In this ablation study, we set \(\alpha\) to 1 by default and adjusted \(\beta\) to achieve different influence levels of \( loss_{p-n} \). As illustrated in Figure~\ref{fig:TA-num}(a), we found that as the \(\beta/\alpha\) ratio increases, i.e., the higher the weight of \( loss_{p-n} \), the better defense performance.

\textbf{The pos-/neg- temperature factor}
Since we employ fine-grained alignment of positive and negative subtexts with images, it is essential to consider the relationship between the model's focus on positive and negative samples and the final defense performance. This relationship can be modulated by adjusting the temperature factors \(t_p\) and \(t_n\) in Eq.~\ref{Li2t} and~\ref{Lt2i} to achieve different levels of attention to positive and negative samples. The smaller the temperature factor, the higher the attention received. As shown in Figure~\ref{fig:TA-num}(b), we conducted ablation experiments with five different sets of temperature factors and found that when \(t_p\) is higher than \(t_n\), the model de-emphasizes negative samples, leading to an inability to fine-tune the distribution of text features in the feature space, thus failing to actively distance itself from poisoned image features and resulting in poorer defense performance. Furthermore, if both factors are the same and relatively large, the fine-grained optimization weight of the model decreases, leading to the defense performance decrea. We found that a \(t_n\) of 0.3 yields strong defense performance, and when \(t_p = 0.3\), the impact on BA is minimal. Therefore, we ultimately adopt \(t_p = t_n = 0.3\) as our default setting.

\textbf{The number of samples in every epoch applied CleanerCLIP}
Furthermore, since we do not perform text augmentation on all samples, but rather randomly select a subset of samples to implement CleanerCLIP in each iteration, we explored the impact of sample quantity, as illustrated in Figure~\ref{fig:TA-num}(c). It can be observed that for simpler attacks like BadNet, only 50 samples are sufficient to significantly reduce the ASR after the first iteration, achieving extremely fast and optimal defense performance. For more complex new attack techniques, such as BadCLIP, only 2000 to 3000 samples are needed. This represents a very small training cost compared to the scale of the fine-tuning dataset (500K). 
Regarding model utility, we observe that increasing the number of neg-samples does not notably impact BA: when the num is 50, 250, 500, 750, 1000, 2000, and 3000, the Top-1 BA (\%) are 51.53, 51.57, 51.51, 51.46, 51.47, 51.32, and 51.29, respectively. These results show that the increase in negative samples significantly reduces ASR while maintaining model utility, thanks to the stabilizing effect of positive sub-texts on clean sample features.

\textbf{More ablation results}
In the supplementary materials, we provide additional experimental results to validate our effectiveness. We present the testing results on other MCL models in Sec 9, such as EVA-CLIP, and multimodal datasets like SBUCaption in Sec 10. Additionally, we provide a detailed comparison of the performance with CleanCLIP's EDA text augmentation method in Sec 11, conducting self-supervised fine-tuning for each augmentation strategy to evaluate its defensive capabilities and compare them with our approach. In Sec 12, we also utilize existing open-source text augmentation strategies, like DeCLUTR, as a substitute for our counterfactual semantic enhancement component and compare their performance in fine-tuning defense.

\section{Conclusion and limitation}
In this paper, we focus on fine-tuning defense strategies against backdoor attacks targeting MCL. We propose CleanerCLIP, a counterfactual semantic enhancement method that effectively defends against backdoor attacks in multi-modal contrastive learning models. CleanerCLIP achieves superior defense performance across various datasets and attack scenarios, significantly reducing ASR while maintaining benign accuracy.

\textbf{Limitations} Since our proposed CleanerCLIP primarily addresses backdoor attacks in the image modality, the defense performance against text modality attacks remains unknown. In the future, we will further explore comprehensive and efficient defense methods that are effective across various modalities.

{
    \small
    \bibliographystyle{ieeenat_fullname}
    \bibliography{main}
}

\end{document}